\documentclass[times,twocolumn,final,numbers]{elsarticle}

\usepackage{prletters}
\usepackage{framed,multirow}

\usepackage{amssymb}
\usepackage[fleqn]{amsmath}
\usepackage{latexsym}

\usepackage[nodots]{numcompress}

\usepackage{url}
\usepackage[table,xcdraw]{xcolor}
\usepackage{xcolor}
\definecolor{newcolor}{rgb}{.8,.349,.1}
\usepackage{balance}
\usepackage{graphicx}
\usepackage{multirow}
\usepackage[table,xcdraw]{xcolor}

\journal{Pattern Recognition Letters}
\begin{document}

\begin{frontmatter}

\title{Deep Uncalibrated Photometric Stereo via Inter-Intra Image Feature Fusion}

\author[1]{Fangzhou \snm{Gao}} 
\ead{gaofangzhou@tju.edu.com}

\author[1]{Meng \snm{Wang}}
\ead{meng.wang@tju.edu.cn}

\author[1]{Lianghao \snm{Zhang}}
\ead{opoiiuiouiuy@tju.edu.cn}

\author[1]{Li \snm{Wang}}
\ead{li\_wang@tju.edu.cn}

\author[1]{Jiawan \snm{Zhang} \corref{cor1}}
\cortext[cor1]{Corresponding author.}
\ead{jwzhang@tju.edu.cn}

\address[1]{Department of Intelligence and Computing, Tianjin University, Tianjin, China}

\begin{abstract}
Uncalibrated photometric stereo is proposed to estimate the detailed surface normal from images under varying and unknown lightings. Recently, deep learning brings powerful data priors to this underdetermined problem. This paper presents a new method for deep uncalibrated photometric stereo, which efficiently utilizes the inter-image representation to guide the normal estimation. Previous methods use optimization-based neural inverse rendering or a single size-independent pooling layer to deal with multiple inputs, which are inefficient for utilizing information among input images. Given multi-images under different lighting, we consider the intra-image and inter-image variations highly correlated. Motivated by the correlated variations, we designed an inter-intra image feature fusion module to introduce the inter-image representation into the per-image feature extraction. The extra representation is used to guide the per-image feature extraction and eliminate the ambiguity in normal estimation. We demonstrate the effect of our design on a wide range of samples, especially on dark materials. Our method produces significantly better results than the state-of-the-art methods on both synthetic and real data.
\end{abstract}


\end{frontmatter}


\section{Introduction}
Photometric stereo is proposed to estimate the surface normal from images captured by a fixed camera under varying and known lighting. Compared with other stereo vision methods like multi-view stereo, photometric stereo can produce more detailed normal and perform well on textureless objects. The pioneering photometric stereo method is proposed for the idea Lambertian surface \cite{woodham1980photometric}, following researchers have extended it to handle a wide range of complex surfaces \cite{solomon1996extracting,barsky2003,chandraker2007shadowcuts,mukaigawa2007analysis,verbiest2008photometric,miyazaki2010median,yu2010photometric,wu2010photometric,georghiades2003incorporating,chung2008efficient,wu2010robust,ikehata2012robust}.

However, these methods rely on complex light calibration. To overcome it, uncalibrated photometric stereo is proposed to accomplish the task without light calibration. To reduce the ill-posedness caused by the lack of lighting information, most traditional methods assume an ideal Lambertian surface \cite{alldrin2007resolving,papadhimitri2014closed,shi2010self} or a uniform light distribution \cite{lu2013uncalibrated,lu2015uncalibrated}, which limits the practical application. Recently, motivated the significant advancements made by deep learning in computer vision, some researchers \cite{chen2018ps, chen2019self,chen2020learned,kaya2021uncalibrated} have utilized deep learning to leverage data priors and generalized the uncalibrated photometric stereo to the real complex condition.

\begin{figure}
  \centering
  \includegraphics[width=0.5\textwidth]{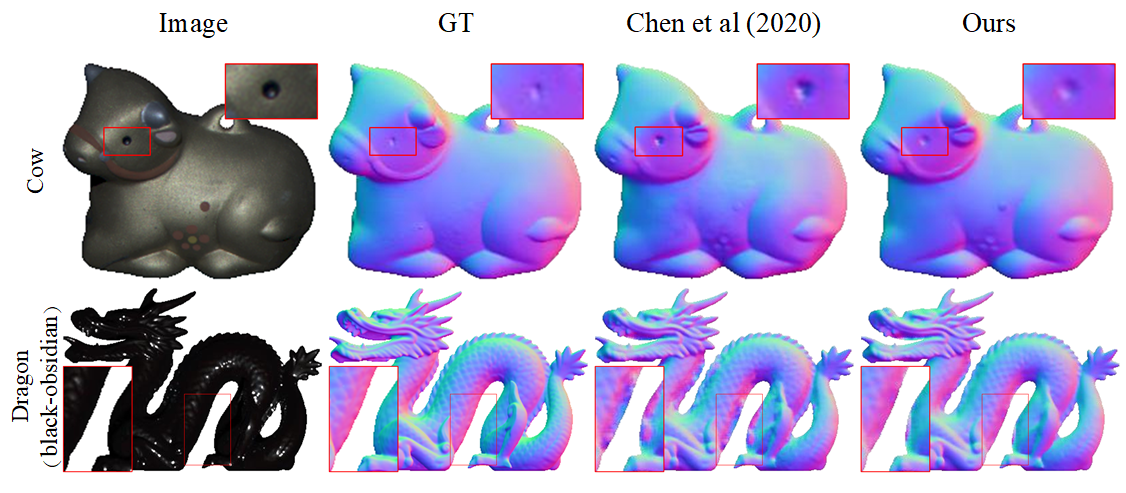}
  \caption{Visualized results of the cow sample in \cite{shi2016benchmark} and the dragon sample in \cite{chen2020learned}, compared with the state-of-the-art method \cite{chen2020learned}. Chen \textit{et al.} \cite{chen2020learned} misjudges the normals on dark materials, especially on regions that lack highlights. In contrast, our method significantly improves the results. The dark regions are marked with red boxes and enlarged.}
\label{fig:tes}
\end{figure}

The main challenge in deep uncalibrated photometric stereo is to enable the network to deal with the unordered and arbitrary numbers of input images. The common CNN-based network is unsuitable since it requires fixed input channels. Some researchers utilized optimization-based neural inverse rendering to solve this problem \cite{kaya2021uncalibrated}. Kaya \textit{et al.} \cite{kaya2021uncalibrated} optimized the surface normal by the neural rendering layers. They explicitly modeled the effect of interreflection and did not rely on the ground-truth of surface normals for training. While their method was limited by the assumption of the differentiable surface and performed poorly on complicated objects. Other researchers used the size-independent pooling layer to aggregate features from different inputs \cite{chen2019self,chen2020learned}. Chen \textit{et al.} \cite{chen2019self} proposed a light calibration network to estimation the lighting for the following normal estimation. They first used a shared-weight feature extractor to explore intra-image variation from each input independently and then fused them using a max-pooling layer to explore the inter-image variation. Chen \textit{et al.} \cite{chen2020learned} further designed a cyclic network structure to introduce extra inter-image guidance and intra-image guidance for the lighting estimation. It improved the accuracy in lighting estimation. However, the single pooling layer was weak in exploring input images' information, which caused the estimated surface normals were still not satisfactory, especially on dark materials, as shown in Fig.\;\ref{fig:tes}.

In this paper, we consider the correlation between intra-image intensity variation and inter-image lighting variation and propose an inter-intra image feature fusion module to combine these two kinds of variations. Specifically, we explore the intra-image variation from each input by a share-weight CNN-based feature extractor that contains our feature fusion modules. During per-image local feature extraction, these fusion modules aggregate global features (like material and rough geometry) among multi-images and introduce them into local feature extraction. The implicit material and geometry representations in global features can guide the next local feature extraction, which allows a more efficient feature extraction and a more accurate normal estimation. Experiments demonstrate that our design significantly improved the results, especially on the challenging dark materials. 

\section{Related Work}
\subsection{Deep Uncalibrated Photometric Stereo}
    Most traditional methods in uncalibrated photometric stereo rely on unpractical assumptions to like an idea Lambertian reflectance model \cite{woodham1980photometric,kumar2017monocular,schonberger2016structure} or a uniform light distribution \cite{furukawa2009accurate,kumar2019superpixel}. On the contrast, the learning-based methods leveraged powerful data priors and performed better on real objects.
    
    To enable the network handle arbitrary numbers of input images in uncalibrated photometric stereo, some researchers utilized neural inverse rendering \cite{kaya2021uncalibrated}, while others utilized size-independent pooling layers to fuse features. 
    
    Kaya \textit{et al.} \cite{kaya2021uncalibrated} calculated the surface normals, BRDFs, and depth by the optimization of neural rendering loss, which explicitly modeled the interreflections. While their neural rendering relied on a continuous surface to compute depth and interreflection kernel, and performed poorly on complex surfaces. Chen \textit{et al.} \cite{chen2018ps} directly predicted surface normal from input images. They used a shared-weight extractor to extract local features from each input before fusing them using a max-pooling layers. Chen \textit{et al.} \cite{chen2019self} further introduced lightings as extra supervision. They first estimated the light directions and intensities, then predicted the surface normral with estimated lightings. Recently, Chen \textit{et al.} \cite{chen2020learned} designed a cyclic network structure for lighting estimation. They first estimated rough lightings and rough surface normal, then provided computed shading and rough normal as extra  guidance for the final lighting estimation. But they only focus on improving accuarcy in lighting estimation, the final results of surface normals need to further improved.

\subsection{Multi-Image Deep Network}
    Similar with photometric stereo, many other tasks in computer vision and computer graph take a variable number of images as input \cite{choy20163d,wiles2017silnet,deschaintre2019flexible,aittala2018burst,qi2017pointnet}. Choy \textit{et al.} \cite{choy20163d} took images as a squeeze and applied a RNN-based network in multi-view 3D reconstruction. While their architecture was sensitive to the order of inputs and paid less attention to latter images. To overcome it, Wiles \textit{et al.} \cite{wiles2017silnet} used a shared-weight feature extractor to extract local features from each image, then fused them to a fix-sized global feature using a order-independent pooling layer. Similar strategies were also adopted in SVBRDF capture \cite{deschaintre2019flexible}, burst image deblurring \cite{aittala2018burst}, deep learning on 3d points \cite{qi2017pointnet}. Inspired by these works, we designed our network for uncalibrated photometric stereo, which leverages the correlation between images efficiently.

    \begin{figure}
    \centering
    \includegraphics[width=0.45\textwidth]{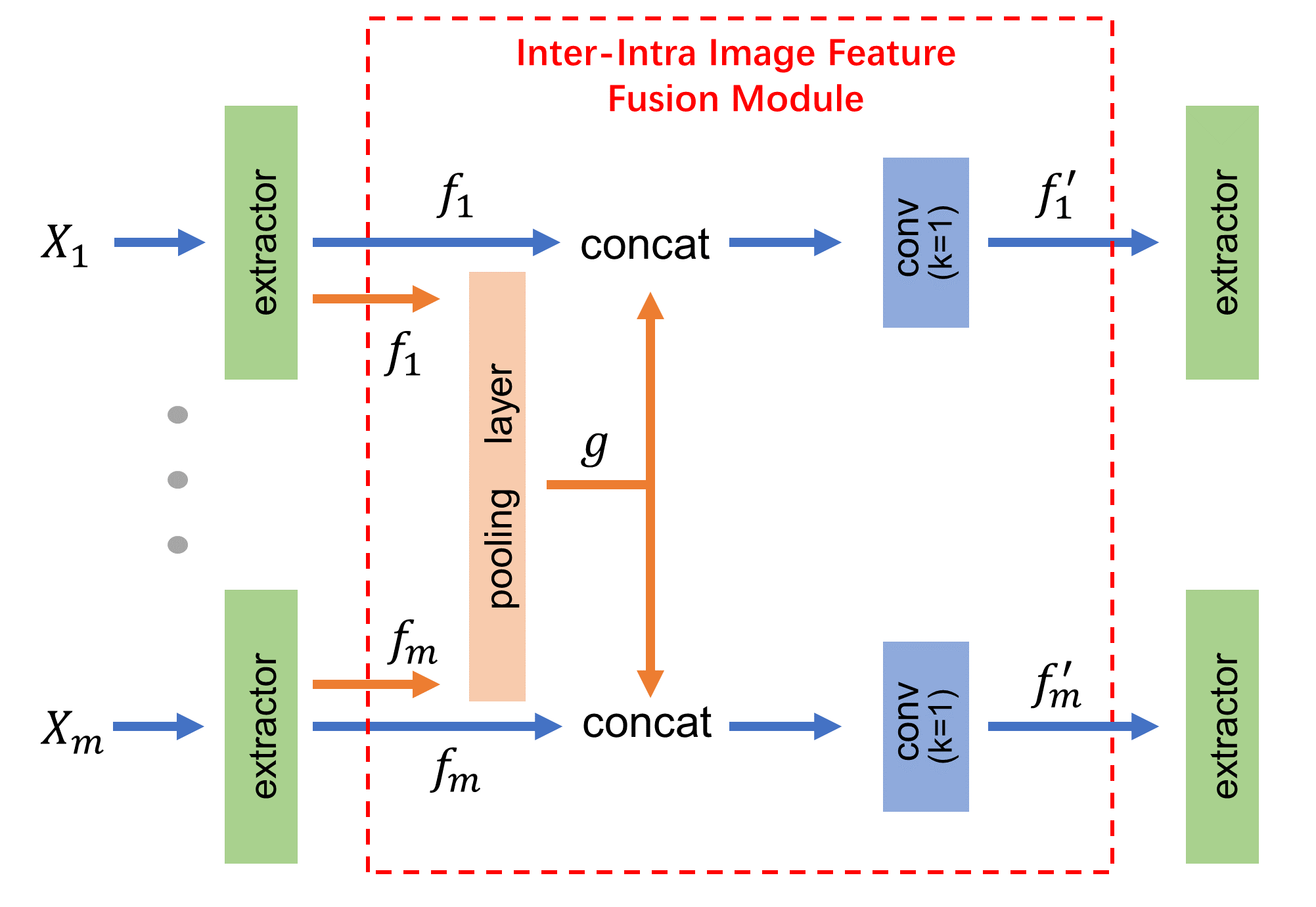}
    \caption{An illustration of the inter-intra image feature fusion module. It aggregates global features from the intermediate local features of each input and concatenates them with per-image local features separately for the following 1 × 1 convolutional layer, which fuses local features with global features. }
    \label{fig:zoom in}
    \end{figure}
    
    \begin{figure*}
    \centering
    \includegraphics[width = 0.95\textwidth]{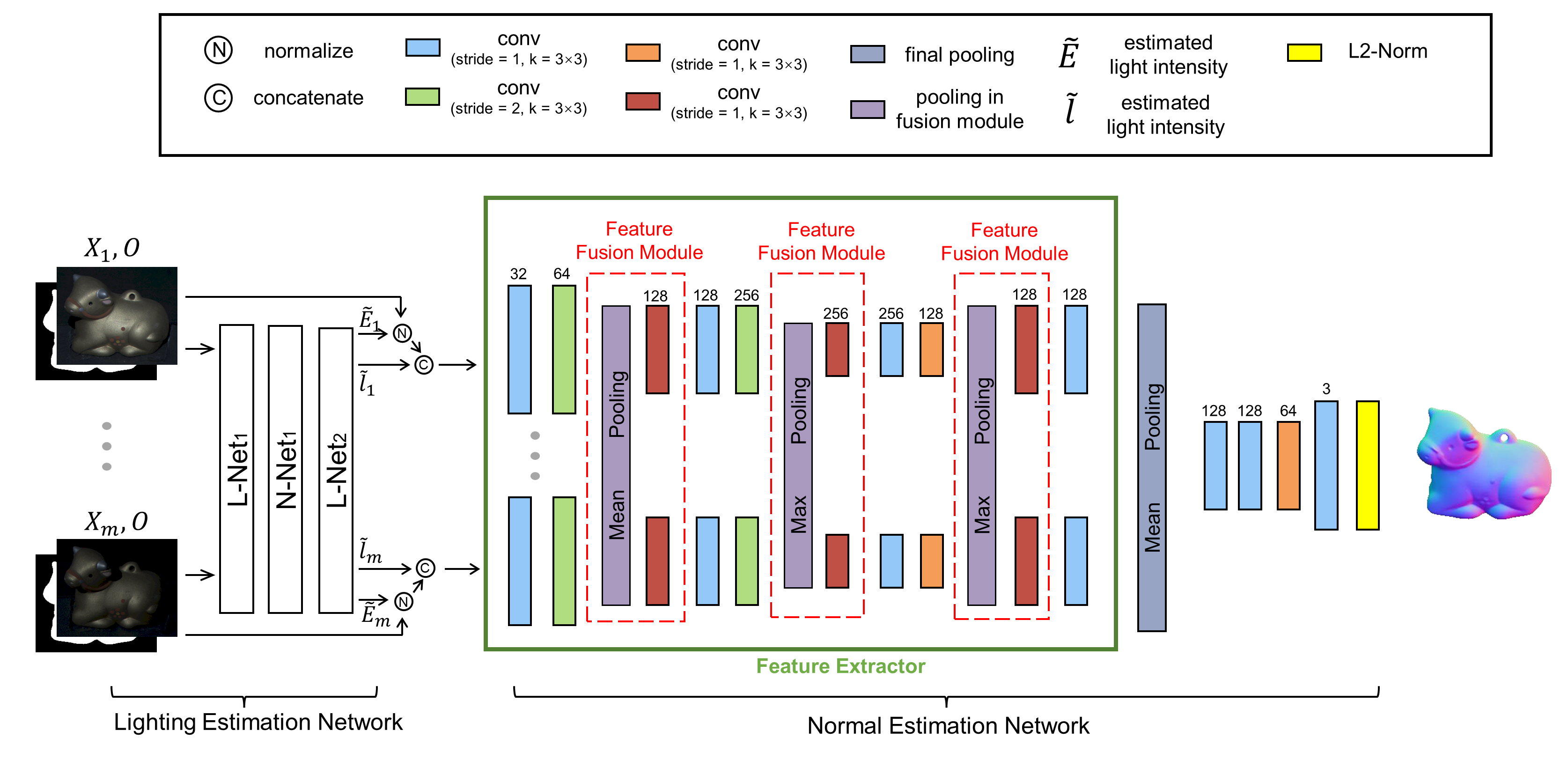}
    \caption{Overview of our framework. With our feature fusion module, our normal estimation network combines inter-image features with local features during local feature extraction. }
    \label{fig:network}
    \end{figure*} 
    
\section{Our Method}
This section firstly introduces our motivation and strategy of exploring the intra-image and inter-image variations and then presents our network structure. 

Following the common assumptions, we assume that images are captured by a radiometrically calibrated orthogonal camera under single directional lighting. Moreover, we use ``intensity" to refer to image irradiance for simplicity.

    \subsection{Inter-Intra Image Feature Fusion Module}
    \label{combination}
    In deep uncalibrated photometric stereo, the intra-image intensity variation and inter-image lighting variation of input images are correlated and significant for normal estimation. The inter-image variation under different light conditions implies the surface material and geometry information, which is crucial for eliminating the ambiguity of the normal, light and reflectance model. A common approach to exploring these two kinds of variation is to use a shared-weight extractor to extract per-image features independently from each input, then fuse them using a pooling layer \cite{wiles2017silnet,chen2018ps,deschaintre2019flexible}. However, this network structure can not perceive any inter-image variation to eliminate the ambiguity in per-image feature extraction.
    
    To over this problem, we propose the inter-intra image feature fusion module. Specifically, For a set of input images $X = [X_1,X_2, ...,X_m]$ a set of $M$ input images, there are a set of per-image features $f = [f_1,f_2, ...,f_m]$ extracted by the front layers in the share-weight extractor. The modules inserted fuses features as follow:
    \begin{equation}
        g=Pool\left(f_{1}, f_{2}, \cdots \cdots ,f_{m}\right),
    \end{equation}
    where ${Pool}$ represents the pooling layer that aggregates global features ${g}$ from intermediate local features $f$.
    \begin{equation}
    f_{i}^{\prime}=N_{f u s e}\left(f_{i}, g\right),
    \end{equation}
    where ${N_{f u s e}}$ represents the 1×1 convolution layer that fuses local features with global features separately to obtain the new per-image features $f_{i}^{\prime}$. The new per-image features $f^{\prime} = [f_{1}^{\prime},f_{2}^{\prime}, ...,f_{m}^{\prime}]$ are fed to the rest layers in the extractor, as shown in Figure\;\ref{fig:zoom in}.  
    
    With the global features representing the inter-image variation of all images, the extractor can utilize the implied material and geometry features to extract local features more accurately and efficiently. For instance, it is easy for the extractor to roughly distinguish the shadows and regions with low albedo since the intensities of common shadows change rapidly with the changing lighting while the intensities of regions with low albedo remain low. Moreover, with the extra cues of other inputs, the network can capture the slight changes of the intensities of dark materials, which provide strong cues for inferring the surface normal.

    \subsection{Network Structure}
    Our method contains two networks: the lighting estimation network and the normal estimation network. Given $X = [X_1,X_2, ...,X_m]$ a set of $M$ input images and the object mask O, as shown in Figure\;\ref{fig:network}, we first estimate the light directions and intensities for each image using the lighting estimation network. Then the estimated lighting is used to recover the surface normal using our proposed normal estimation network.
    
    \textbf{Lighting Estimation Network} For the lighting estimation network, we follow the structure proposed in \cite{chen2020learned}. As shown in Figure\;\ref{fig:network}, the lighting estimation network contains three sub-networks, including two lighting estimation sub-networks (L-Net$_1$ and L-Net$_2$) and a normal estimation sub-network (N-Net). The L-Net$_1$ estimates initial lighting given the input images and object mask. Then the N-Net predicts surface normal given the initial lighting and input images. Finally, the L-Net$_2$ estimates the final lighting with extra rough normal and shading estimated by the front networks. More details can be found in \cite{chen2020learned}.
    
    \textbf{Normal Estimation Network} 
    With the estimated lighting, we normalize input images with the corresponding predicted light intensity, and then concatenate them with the corresponding predicted light direction as the inputs. As shown in Figure\;\ref{fig:network}, for a set of image-lighting pairs, per-image local features are extracted separately by a shared-weight feature extractor before fusing them to global features. Then the following several convolutional layers and an L2-normalization layer further infer the normal map from global features.
    
    The CNN-based feature extractor is plugged with the inter-intra image feature fusion modules at multiple levels. During the local feature extraction, as motioned in section\;\ref{combination}, the feature fusion modules introduce the global features among multi-images to guide the per-image feature extraction. We choose the mean-pooling layer for the first fusion block to obtain the material and reflectance representations and the max-pooling layer for others.
    
\begin{figure}[t]
    \centering
    \includegraphics[width=0.475\textwidth]{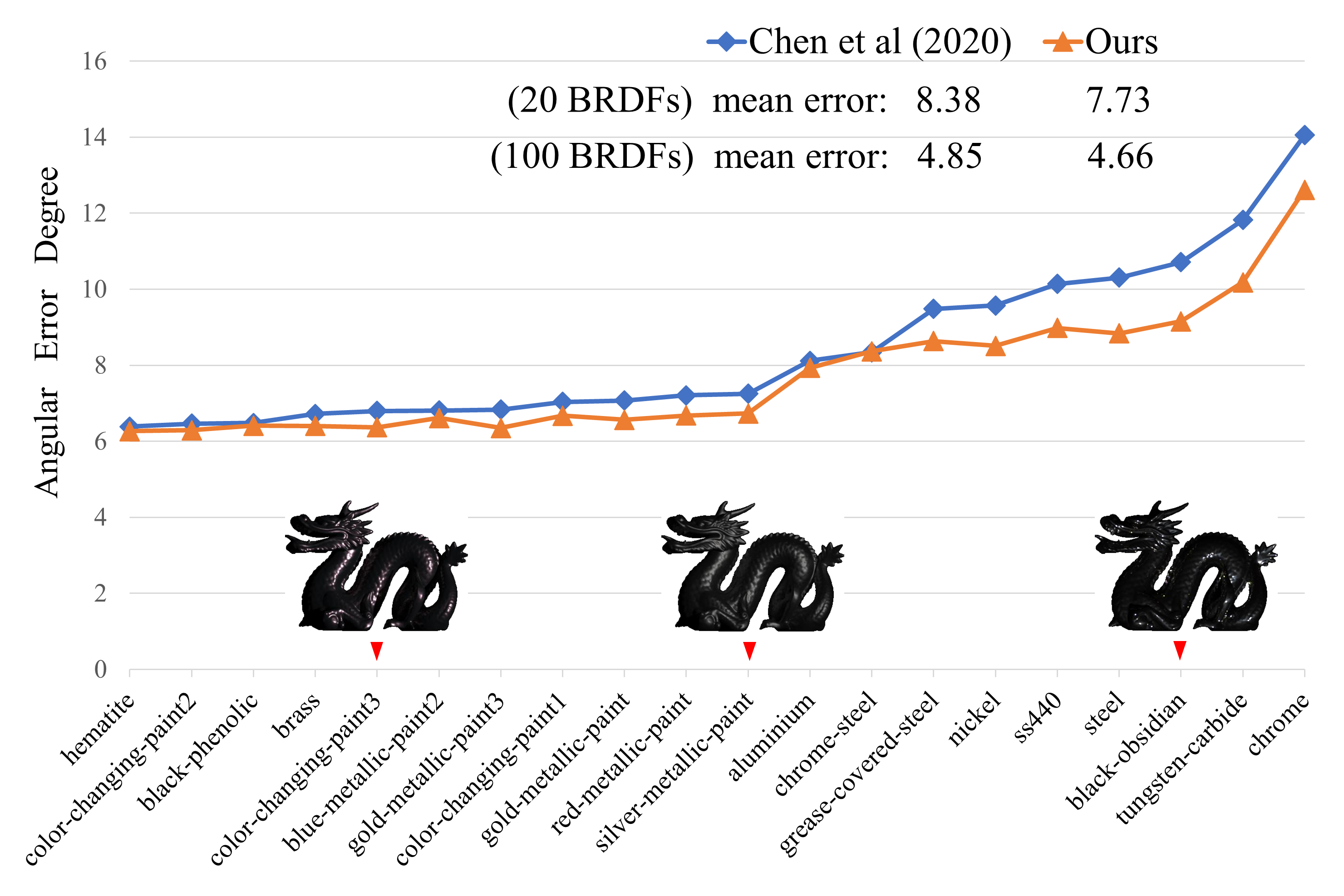}
    \caption{Quantitative results of of dragon test data in \cite{chen2020learned}, compared with Chen \textit{et al.} \cite{chen2020learned}. Our method has lower mean error for both 20 BRDFs shown in the figure and all 100 BRDFs. Besides, our method performs significantly better for those dark materials which are challenging for Chen's method. Images above the horizontal axis show the corresponding samples}
    \label{fig:syn_compare}
\end{figure}

    \subsection{Loss Function}
    The lightings are discretized and considered as a classification problem (32 classes for elevation and azimuth to represent light direction, 32 classes for intensity). Given ${M}$ images, the loss function for L-Net in lighting estimation network is 
    \begin{equation}
        \mathcal{L}_{\text {light }}=\frac{1}{M} \sum_{f}\left(\mathcal{L}_{l_{a}}^{m}+\mathcal{L}_{l_{e}}^{m}+\mathcal{L}_{e}^{m}\right),
    \end{equation}
    where ${\mathcal{L}_{l_{a}}^{m}}$,${\mathcal{L}_{l_{e}}^{m}}$ and ${\mathcal{L}_{e}^{m}}$ are the cross-entropy loss for light azimuth, elevation, and intensity classifications, respectively.
    
    The normal loss function of the normal estimation network and the N-Net in lighting estimation network is
    \begin{equation}
        \mathcal{L}_{\text {normal }}=\frac{1}{P} \sum_{p}\left(1-\boldsymbol{n}_{p}^{\top} \tilde{\boldsymbol{n}}_{p}\right),
    \end{equation}
    where ${P}$ donates the number of pixels in per image, and ${\boldsymbol{n}_{p}}$ and ${\tilde{\boldsymbol{n}}_{p}}$ are the ground truth and predicted normal at pixel $p$, respectively.
    And we fine-tune the entire lighting estimation network end-to-end using the following loss:
    \begin{equation}
        \mathcal{L}_{\text {fine-tune }}=\mathcal{L}_{\text {light}_{1}}+\mathcal{L}_{\text {normal }}+\mathcal{L}_{\text {shading}}+\mathcal{L}_{\text {light}_{2}},
    \end{equation}
    
    \begin{equation}
        \mathcal{L}_{\text {shading }}=\frac{1}{M P} \sum_{m} \sum_{p}\left(\boldsymbol{n}_{p}^{\top} \boldsymbol{l}_{m}-\tilde{\boldsymbol{n}}_{p}^{\top} \tilde{\boldsymbol{l}}_{m}\right)^{2},
    \end{equation}
    where $\mathcal{L}_{\text {shading}}$ denote $\mathcal{L}_{\text{light}_{1}}$ and $\mathcal{L}_{\text {light}_{2}}$ denote the loss function in L-Net$_1$ and L-Net$_2$, $\mathcal{L}_{\text {shading}}$ denote the shading loss, and $\boldsymbol{l}_{m}$ and $\tilde{\boldsymbol{l}}_{m}$ are the ground truth and predicted light direction for the ${m}^{th}$ image.

    \begin{figure}
    \centering
    \includegraphics[width=0.48\textwidth]{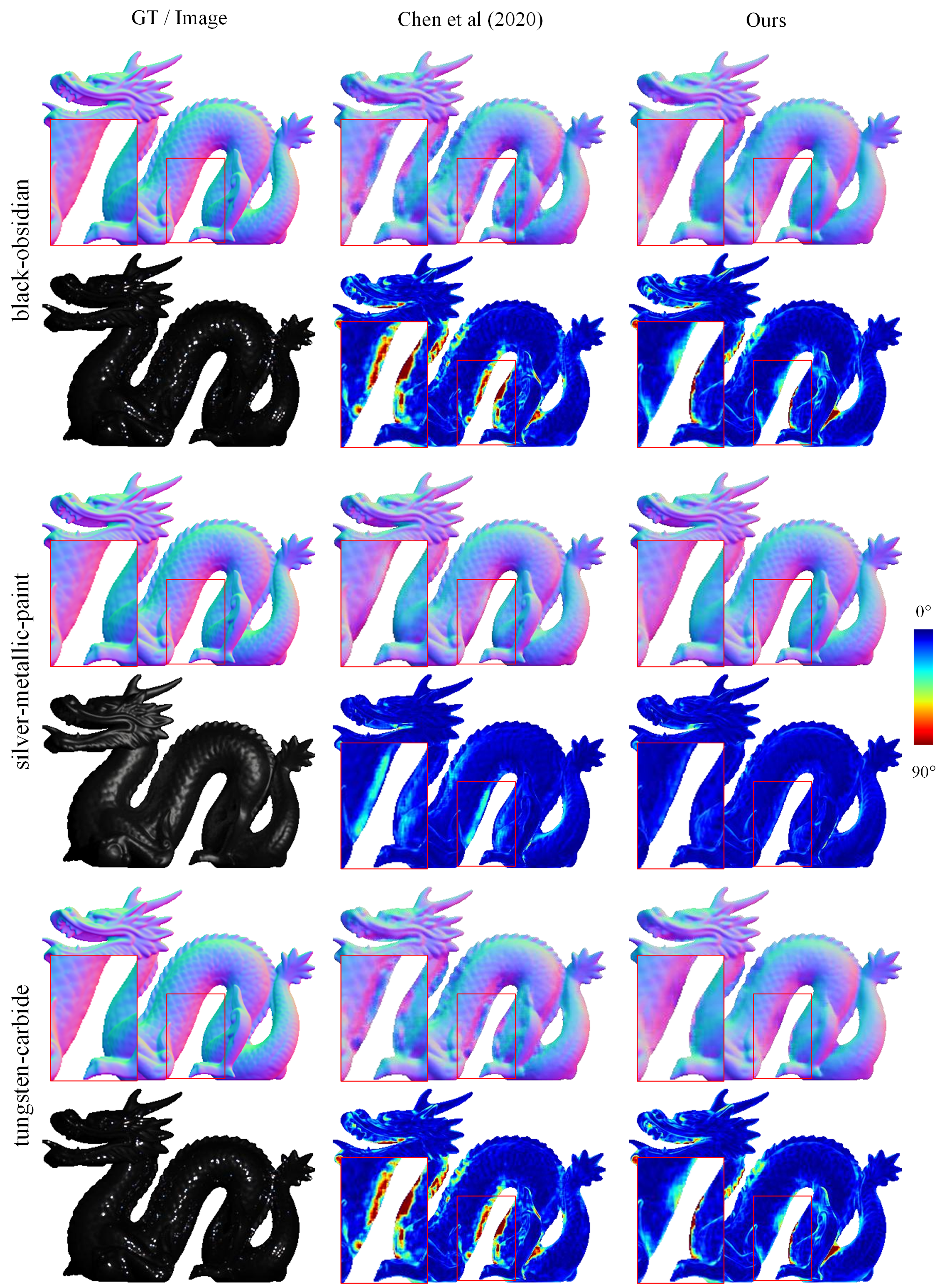}
    \caption{Visualized results of dragon test data in \cite{chen2020learned}, compared with Chen \textit{et al.} \cite{chen2020learned}. The visualized error maps lie underneath the estimated normal maps. The regions where lacks highlight due to self-occlusion are marked with red boxes and enlarged at the bottom-left corner.  }
    \label{fig:syn_visual}
    \end{figure}

\begin{table*}[t]
        
\centering
\caption{Quantitative results on DiLiGenT benchmark \cite{shi2016benchmark}, compared with comparison with traditional and deep uncalibrated photometric stereo methods. For each object, the best result is bolded and colored in dark-red, and the second best result is colored in light-red. \dag \ means we use the deeper vision of version of UPS-FCN. \cite{chen2018ps}}
\label{tab:results_dili}
\resizebox{0.95\textwidth}{!}{
\begin{tabular}{cccccccccccc}
\hline
\textbf{Methods}                              & \textbf{Ball}                         & \textbf{Cat}                          & \textbf{Pot1}                          & \textbf{Bear}                         & \textbf{Pot2}                         & \textbf{Buddha}                       & \textbf{Goblet}                       & \textbf{Reading}                       & \textbf{Cow}                          & \textbf{Harvest}                       & \textbf{Average}                      \\ \hline
\cellcolor[HTML]{FFFFFF}Papadh. et al. (2014) \cite{papadhimitri2014closed} & \cellcolor[HTML]{FFFFFF}4.77          & \cellcolor[HTML]{FFFFFF}9.54          & 9.51                                  & 9.07                                  & 15.90                                 & 14.92                                 & 29.93                                 & 24.18                                  & 19.53                                 & 29.21                                  & 16.66                                 \\
\cellcolor[HTML]{FFFFFF}Lu et al. (2017)  \cite{lu2017symps}    & \cellcolor[HTML]{FFFFFF}9.30          & \cellcolor[HTML]{FFFFFF}12.60         & 12.40                                 & 10.90                                 & 15.70                                 & 19.00                                 & 18.30                                 & 22.30                                  & 15.00                                 & 28.00                                  & 16.30                                 \\
\cellcolor[HTML]{FFFFFF}UPS-FCN† (2018) \cite{chen2018ps}       & \cellcolor[HTML]{FFFFFF}3.96          & \cellcolor[HTML]{FFFFFF}12.16         & 11.13                                 & 7.19                                  & 11.11                                 & 13.06                                 & 18.07                                 & 20.46                                  & 11.84                                 & 27.22                                  & 13.62                                 \\
\cellcolor[HTML]{FFFFFF}SDPS-Net (2019) \cite{chen2019self}      & \cellcolor[HTML]{FFCCCC}2.77          & \cellcolor[HTML]{FFFFFF}8.06          & 8.14                                  & 6.89                                  & \cellcolor[HTML]{FFCCCC}7.50          & 8.97                                  & 11.91                                 & \cellcolor[HTML]{FFCCCC}14.90          & 8.48                                  & 17.43                                  & 9.51                                  \\
Chen et al.(2020) \cite{chen2020learned}                  & \cellcolor[HTML]{FF9999}\textbf{2.48} & \cellcolor[HTML]{FFCCCC}7.87          & \cellcolor[HTML]{FF9999}\textbf{7.21} & \cellcolor[HTML]{FFCCCC}5.55          & \cellcolor[HTML]{FF9999}\textbf{7.05} & \cellcolor[HTML]{FFCCCC}8.58          & \cellcolor[HTML]{FFCCCC}9.62          & 14.92                                  & \cellcolor[HTML]{FFCCCC}7.81          & \cellcolor[HTML]{FFCCCC}16.22          & \cellcolor[HTML]{FFCCCC}8.73          \\
Kaya et al. (2021) \cite{kaya2021uncalibrated}                            & 3.78                                  & 7.91                                  & 8.75                                  & 5.96                                  & 10.17                                 & 13.14                                 & 11.94                                 & 18.22                                  & 10.85                                 & 25.49                                  & 11.62                                 \\ \hline
Ours                                          & 3.04                                  & \cellcolor[HTML]{FF9999}\textbf{7.55} & \cellcolor[HTML]{FFCCCC}7.54          & \cellcolor[HTML]{FF9999}\textbf{5.40} & 8.05                                  & \cellcolor[HTML]{FF9999}\textbf{8.39} & \cellcolor[HTML]{FF9999}\textbf{8.91} & \cellcolor[HTML]{FF9999}\textbf{14.81} & \cellcolor[HTML]{FF9999}\textbf{6.88} & \cellcolor[HTML]{FF9999}\textbf{15.23} & \cellcolor[HTML]{FF9999}\textbf{8.58} \\ \hline
\end{tabular}}
\end{table*}

    \begin{figure*}
    \centering
    \includegraphics[width=0.95\textwidth]{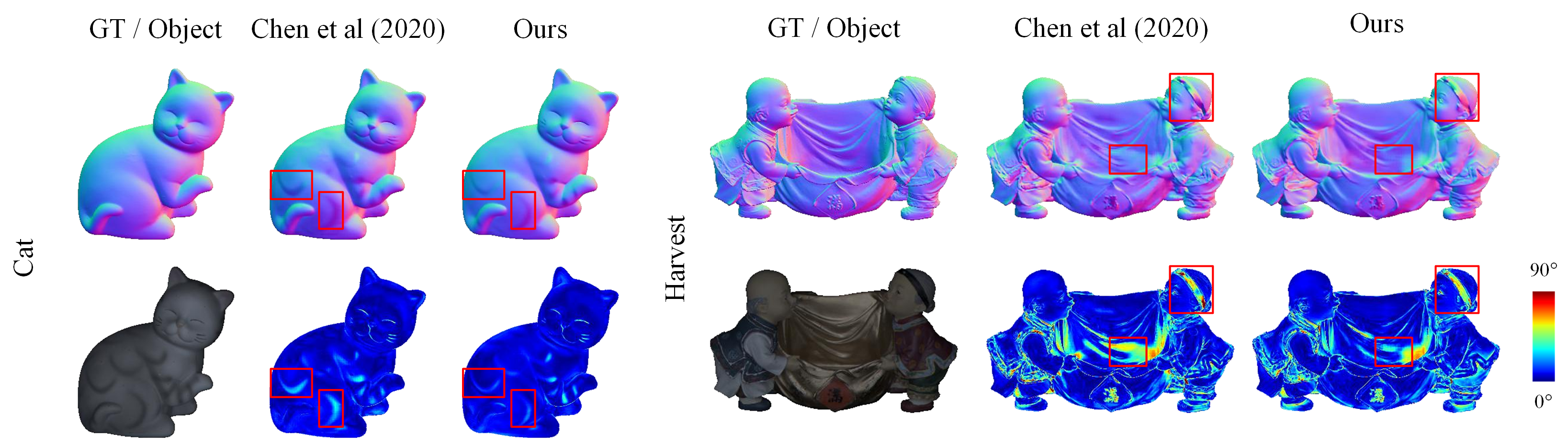}
    \caption{Visualized results of DiLiGenT benchmark \cite{shi2016benchmark}, compared with Chen \textit{et al.} \cite{chen2020learned}. The object image is produced by averaging all images for better visualization. The visualized error maps lie underneath the estimated normal maps. The red boxes demonstrate our improvement in dark materials and concave regions. The full results of our method on DiLiGenT benchmark are included in the supplementary.  }
    \label{fig:dili_compare}
    \end{figure*}

\begin{figure}[t]
    \centering
    \includegraphics[width=0.45\textwidth]{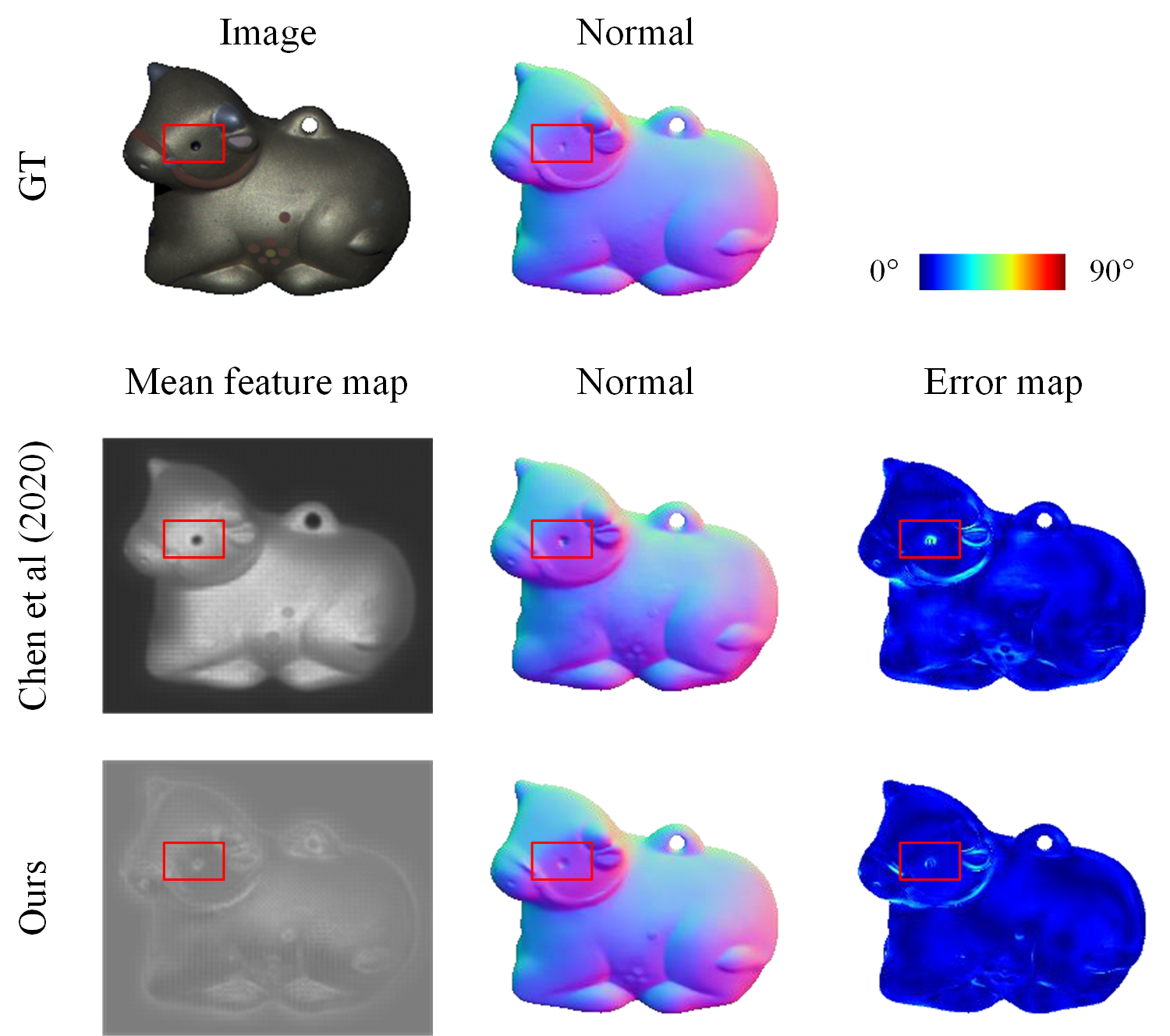}
    \caption{Visualized results on the feature domain, compared with Chen \textit{et al.} \cite{chen2020learned}. The mean feature maps are multiplied by 5 for better visualization. The lower intensity of our mean feature map is because we used a mean-pooling layer for the final feature fusion while Chen \textit{et al.} used a max-pooling layer. The cow's dark eye is marked with red boxes.}
    \label{fig:feature_compare}
\end{figure}

\section{Experiment}

We evaluated and analyzed our method on synthetic and real data and used the popular mean angular error (MAE) to measure the error of the estimated normal.

\subsection{implement details}

We trained our model on the publicly available synthetic Blobby and Sculpture Dataset \cite{chen2018ps}, which contains 85,212 surfaces and each is illuminated under 64 random light directions. 

For the lighting estimation, we followed the training procedure in \cite{chen2020learned} to train three sub-networks one after another until convergence, then fine-tuned the lighting estimation network end-to-end. The normal estimation network was trained with ground truth lighting and surface normal, with a batch size of 32 for 30 epochs. We used the same training configuration in \cite{chen2020learned}, including the learning rate, batch size, number of training epochs, etc, to further demonstrate the superiority of our network architecture.

We implemented the framework in PyTorch and used the Adam optimizer \cite{kingma2014adam} with default parameters. It took 4.63 hours to train the normal estimation network with a 3.70GHz Intel Core i9 CPU and a single NVIDIA GeForce RTX 3060 GPU.

\subsection{Evaluation on Synthetic Data}
    We compared our method with the state-of-the-art method \cite{chen2020learned} on the synthetic dragon dataset in \cite{chen2020learned}. The dragon shape is rendered with 100 MERL BRDFs \cite{matusik2003data} and each is illuminated under 82 randomly sampled light directions. 
    
    In Figure\;\ref{fig:syn_compare}, we show quantitative results of 20 BRDFs on which Chen's method \cite{chen2020learned} performed worst and sort them by the error. Our method produced better results, especially for those challenging dark materials, as we analyzed in section \ref{combination}. Figure\;\ref{fig:syn_visual} shows the visualized results of several dark materials. For marginal regions where lack highlights as extra cues, Chen's method produced unreasonable surface normals, while our method significantly improved them on various materials. It proves the superiority of our method. With the feature fusion modules, our model can infer the surface normal through slight intensities changes among images.

\subsection{Evaluation on Real Data}
    We evaluated our method on the public DiLiGenT benchmark \cite{shi2016benchmark} and report the quantitative results compared with other uncalibrated photometric stereo methods. As shown in Table\;\ref{tab:results_dili}, our method achieved the best performance with the lowest average error and the lowest error for most objects. The visualized results in Figure\;\ref{fig:dili_compare} and Figure\;\ref{fig:tes} also proves that our method significantly improve the normal estimation on dark and concave regions.
    
    Besides, we demonstrated the superiority of our method on the feature domain. We compared our method with Chen \textit{et al.} \cite{chen2020learned}, in which the extractor dose not perceive any inter-image information. We averaged all channels of the fused global features aggregated by the final pooling layer (each channel was normalized). As shown in Figure\;\ref{fig:feature_compare}, few valid features were extracted for dark regions in Chen's method, which explained its poor result of surface normal. While our method extracted valid features for all regions and inferred more accurate normals.

\begin{table*}
\centering
\caption{Quantitative results of the ablation study. For the DiLiGent benchmark, We report the error of all objects and the average error. For Synthetic dragon data, we report the errors of samples of ten typical materials and the average error of all 100 BRDFs. The lowest errors are bolded.}
\label{tab:ablation}
\resizebox{0.95\textwidth}{!}{\
\begin{tabular}{c|ccccccccccc}
\hline
\multirow{2}{*}{Methods} & \multicolumn{11}{c}{DiLiGenT benchmark}                                                                                                                                                                 \\ \cline{2-12} 
                         & Ball          & Cat           & Pot1           & Bear          & Pot2          & Buddha        & Goblet        & Reading        & Cow           & \multicolumn{1}{c|}{Harvest}          & Average       \\ \hline
Ours                     & 3.04          & \textbf{7.55} & \textbf{7.54}  & 5.40          & \textbf{8.05} & \textbf{8.39} & \textbf{8.91} & \textbf{14.81} & \textbf{6.88} & \multicolumn{1}{c|}{\textbf{15.23}}   & \textbf{8.58} \\
Ours v$_1$                  & \textbf{2.85} & 9.1           & 7.86           & \textbf{5.35} & 8.75          & 8.84          & 9.42          & 15.63          & 7.56          & \multicolumn{1}{c|}{16.45}            & 9.18          \\
Ours v$_2$                  & 3.7           & 8.61          & 7.63           & 6.17          & 9.20          & 8.44          & 9.45          & 15.11          & 7.61          & \multicolumn{1}{c|}{16.23}            & 9.21          \\ \hline
\multirow{2}{*}{Methods} & \multicolumn{11}{c}{Synthetic dragon    data}                                                                                                                                                           \\ \cline{2-12} 
                         & Aluminium     & Blue-acrylic  & Chrome         & Delrin        & Nickel        & Nylon         & PVC           & SS440          & Steel         & \multicolumn{1}{c|}{Tungsten-Carbide} & Average       \\ \hline
Ours                     & \textbf{7.93} & \textbf{3.79} & \textbf{12.62} & \textbf{3.25} & \textbf{8.51} & \textbf{3.85} & 3.59          & \textbf{8.98}  & \textbf{8.84} & \multicolumn{1}{c|}{\textbf{10.18}}   & \textbf{4.66} \\
Ours v$_1$                  & 8.57          & 4.65          & 13.91          & 3.55          & 9.88          & 4.69          & \textbf{3.58} & 10.14          & 10.20         & \multicolumn{1}{c|}{11.55}            & 5.03          \\
Ours v$_2$                  & 8.33          & 5.21          & 14.01          & 3.36          & 9.58          & 4.80          & 3.95          & 9.74           & 11.20         & \multicolumn{1}{c|}{11.13}            & 5.03          \\ \hline
\end{tabular}}
\end{table*}

\subsection{Ablation Study}
\begin{figure}[h]
    \centering
    \includegraphics[width=0.45\textwidth]{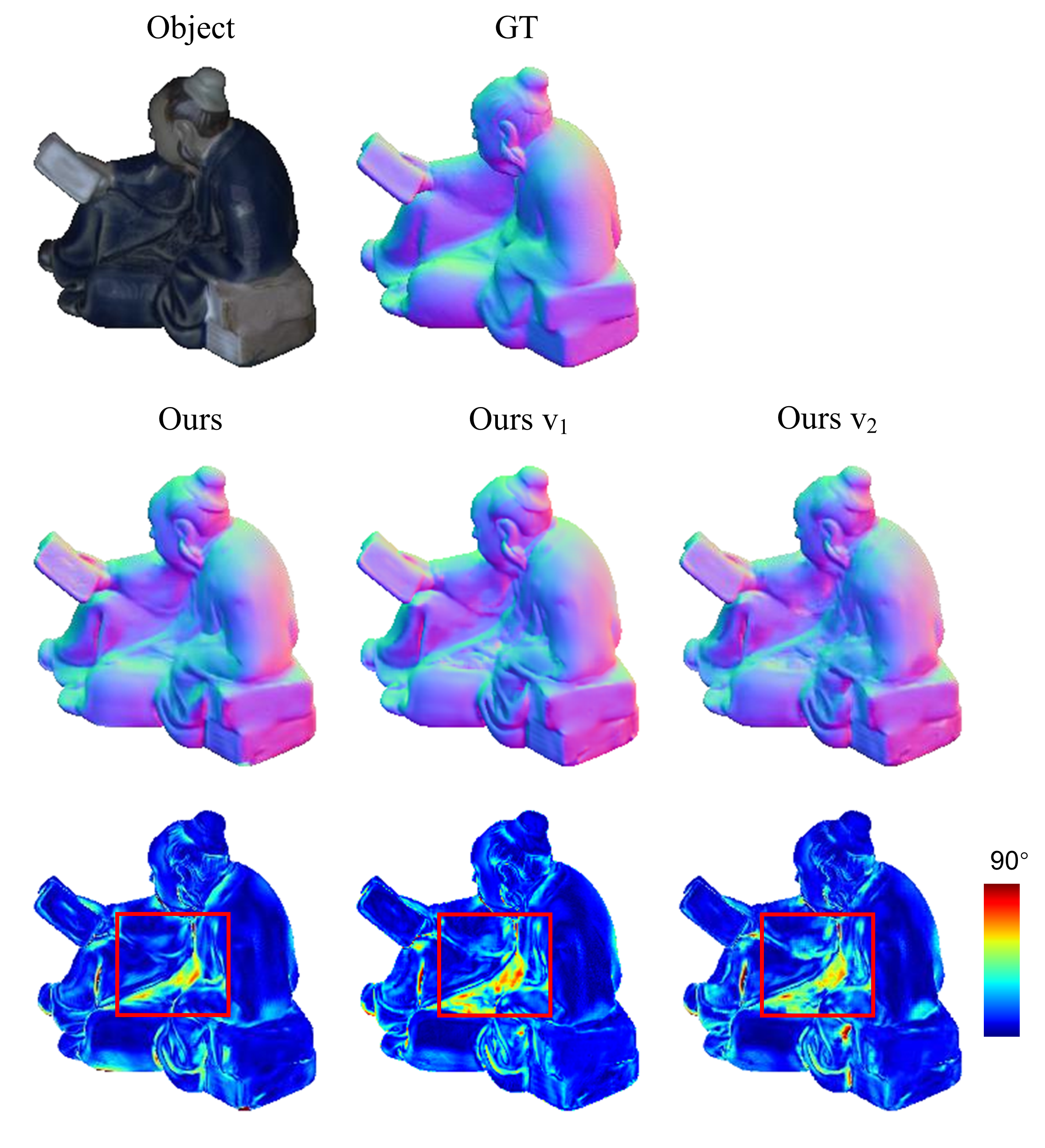}
    \caption{Visualized results of the ablation study. The object is the ``Reading" from DiLiGenT benchmark \cite{shi2016benchmark}. The concave regions are marked with red boxes.}
    \label{fig:ablation_visual}
\end{figure}

    To validate the effect of perceiving inter-image variation during local feature extraction, we removed all feature fusion modules, then trained our network from scratch in the same configuration. And we also evaluated the model that only be removed the pooling layers in feature fusion modules to prove our method does not simply benefit from a deeper network.  The quantitative results of DiLiGenT benchmark \cite{shi2016benchmark} and dragon test data in \cite{chen2020learned} are summarized in Table\;\ref{tab:ablation}. The visualized results are shown in Figure\;\ref{fig:ablation_visual}. Ours v$_1$ represents the model only without pooling layers in modules. Ours v$_2$ represents the model without all feature fusion modules. 
    
    As shown in Table\;\ref{tab:ablation}, our full method has the lowest errors for almost all  of the samples and the lowest average errors on both real and synthetic data. The visualized results in Figure\;\ref{fig:ablation_visual} also prove the improvement on concave regions. It is clearly shown that the feature fusion modules have an important effect on results, which proves the variation among multi-images can greatly improve the per-image feature extraction.

\section{Conclusion}
This paper proposed the inter-intra image feature fusion module for uncalibrated photometric stereo. With the proposed feature fusion module, the representations of inter-image variations are utilized to guide the per-image feature extraction, which makes the per-image local feature extraction more accurate and efficient for normal estimation.

The experimental results on the feature domain strongly demonstrate the effectiveness of our proposed feature fusion module. In addition, the quantitative and qualitative results show that our method performs significantly better on dark materials than the state-of-the-art method. 

\section*{Declaration of Competing Interest}
The authors declare that they have no known competing financial interests or personal relationships that could have appeared to influence the work reported in this paper.

\section*{Acknowledgments}
This work was supported in part by the National Key Research and Development Program of China (Grant No.2019YFC1521200) and in part by the National Natural Science Foundation of China, No. 62172295.

\bibliographystyle{elsarticle-num}

\end{document}